\documentclass{article}

\usepackage[final, nonatbib]{neurips_2019}

\usepackage[utf8]{inputenc} 
\usepackage[T1]{fontenc}    
\usepackage{hyperref}       
\usepackage{url}            
\usepackage{booktabs}       
\usepackage{amsfonts}       
\usepackage{nicefrac}       
\usepackage{microtype}      
\usepackage{subfigure}
\usepackage{graphicx}
\usepackage{multicol} 
\usepackage{url}
\usepackage{amsmath}
\usepackage{caption}
\usepackage{color}
\usepackage{float}
\usepackage{listings}
\usepackage{amsmath}
\usepackage{amssymb}
\usepackage[utf8]{inputenc}
\usepackage[english]{babel}
\usepackage[normalem]{ulem}
\usepackage{epstopdf}
\usepackage[ruled, lined, longend, linesnumbered]{algorithm2e}
\DeclareGraphicsExtensions{.png,.pdf} 
\DeclareMathOperator{\E}{\mathbb{E}}

\makeatletter
\newcommand{\printfnsymbol}[1]{%
	\textsuperscript{\@fnsymbol{#1}}%
}
\makeatother

\title{CoachNet: An Adversarial Sampling Approach for Reinforcement Learning}

\author{
  Elmira Amirloo Abolfathi\\
  Noah's Arc Lab\\
  Huawei Technologies
  Canada\\
  \texttt{elmira.amirloo@huawei.com} \\
  \And
  Jun Luo\thanks{denotes equal contribution}\\
  Noah's Arc Lab\\
  Huawei Technologies
  Canada\\
  \texttt{jun.luo1@huawei.com} \\
  \And
  Peyman Yadmellat\printfnsymbol{1}\\
  Noah's Arc Lab\\
  Huawei Technologies
  Canada\\
  \texttt{peyman.yadmellat@huawei.com} \\
  \And
  Kasra Rezaee\\
  Noah's Arc Lab\\
  Huawei Technologies
  Canada\\
  \texttt{kasra.rezaee@huawei.com}
}

\begin{document}
\maketitle

\begin{abstract}
Despite the recent successes of reinforcement learning in games and robotics, it is yet to become broadly practical. Sample efficiency and unreliable performance in rare but challenging scenarios are two of the major obstacles. Drawing inspiration from the effectiveness of deliberate practice for achieving expert-level human performance, we propose a new adversarial sampling approach guided by a failure predictor named ``CoachNet''. CoachNet is trained online along with the agent to predict the probability of failure. This probability is then used in a stochastic sampling process to guide the agent to more challenging episodes. This way, instead of wasting time on scenarios that the agent has already mastered, training is focused on the agent's ``weak spots''. We present the design of CoachNet, explain its underlying principles, and empirically demonstrate its effectiveness in improving sample efficiency and test-time robustness in common continuous control tasks.
\end{abstract}


\section{Introduction}
Despite the great success of reinforcement learning (RL) in games and robotics \cite{silver2017mastering}, \cite{OpenAI_dota}, \cite{andrychowicz2018learning}, applying it to real-world problems has remained challenging. Efficiently generating enough amount of meaningful training scenarios and robust performance of the deployed agents especially in presence of rare events are open problems in this area.

The vanilla Monte Carlo approach of randomly sampling states is the standard way of training RL agents \cite{schulman2015trust}, \cite{lillicrap2015continuous}, \cite{haarnoja2018soft}. There are a few problems with this approach.

First, much training is devoted to unnecessary repetition. As an agent starts to learn a task, it predictably fails often. As learning progresses, frequency of failures is expected to drop, but some scenarios could remain challenging for a lot longer. However, because the default approach fails to consider the agent's performance, the RL agent may repeatedly encounter scenarios it already mastered, leading to negligible progress. The RL training consequently becomes unreasonably data hungry.

Second, rare challenges are under-represented in the samples. As training progresses, hard examples become more difficult to find in piles of scenarios that the agent can easily solve. This gives the agent little chance to encounter the rare but informative (for training efficiency) and important (for robustness) failures. In safety critical situations, such as autonomous driving where the cost of a single crash outweighs millions of kilometers of flawless operation, the incommensurately large cost of failure means that there is no tightly bounded reward. Such considerations suggest that exploration guided by the current estimate of expected return tend to be overly neglectful of domain shifts and overly optimistic about absence of critical failures, leading to selection of unreliable solutions for test or deployment \cite{zhao2019investigating}.

Third, the exploration-exploitation tradeoff is left to the RL optimization algorithm alone. While it is reasonable to expect the RL algorithm itself to be smart about the trade-off between exploration and exploitation, it is unreasonable to expect it to do all the work. Intrinsic guidance may only go so far. Smart as we humans are, external help is often needed to push us out of our comfort zone (a.k.a. suboptimal region) so that we can explore towards another level of performance. Architecturally speaking, we could systematically take advantage of the difference between training time and test time to ensure adequate exploration, especially if training is done in simulators or safe environments \cite{andrychowicz2018learning}.

Taking inspiration from the psychology of deliberate practice \cite{ericsson1993role}, we propose \textit{CoachNet} guided sampling method to address the above limitations of the default sampling approach. CoachNet predicts the difficulty of an episode \textit{for a specific version of the RL agent} based on the very first few steps of the episode. Difficulty here is defined as the probability of subsequent failure if the episode is allowed to continue. Predictions from the CoachNet can then be used to probabilistically terminate and reject an episode if it is predicted to be easy, or allow it to finish and contribute to parameter updates of the RL agent if it is predicted to be hard. This way, instead of repeating scenarios that the agent has already mastered, training is deliberately focused on the agent's ``weak spots''. Moreover, failures are negatively correlated with the reward signal. If they are under-represented they potentially can have a catastrophic impact on the robustness of the agent. Increased exposure to these failure cases is expected to enhance robustness of the trained agent. Finally, the CoachNet directs the exploration process extrinsically, i.e. from the outside of the RL optimization. This differentiates it from, for example, the critic in an actor-critic model, which could be used to guide the exploration intrinsically.

In what follows, we present the specifics of the design of CoachNet, discuss its underlying principles, and empirically demonstrate its effectiveness in improving sample efficiency and test-time robustness in common continuous control tasks. The main contributions of our paper are:

\begin{itemize}
	\item Introduction of CoachNet for predicting the agent's performance in an episode, which can be used to guide training;
	\item Empirical demonstration that CoachNet trained with self-labeled data from the RL training process can be used to effectively predict failures;
	\item Proposing an adversarial sampling method based on CoachNet where the RL agent is adaptively exposed to what is difficult for it, a method which can be added to any RL algorithm; and
	\item Empirical evaluation of this method on three continuous locomotion control tasks, demonstrating that with the same number of samples, the agent trained with CoachNet achieves higher average reward and lower number of failures.
\end{itemize}

\section{Related Works}

\textbf{Adversarial Reinforcement Learning} Our approach has an adversarial character because it deliberately directs the agent to scenarios that are more challenging for the agent. While adversarial learning for supervised computer vision models has been extensively researched since \cite{goodfellow2014explaining}, adversarial RL research is still in its early days. The robustness of DDQN~\cite{van2016deep} and DDPG in the presence of adversarial attacks were studied in~\cite{pattanaik2018robust} using search algorithm and SGD to find adversarial states. Then, adversarial noise was added to observations so as to make the policy choose the worst action (lowest Q values). However, this approach requires access to the state generator. In Robust Adversarial Reinforcement Learning (RARL) \cite{pinto2017robust}, a protagonist agent tries to fulfil the goal while an adversary agent tries to disrupt its performance. Fictitious Self-Play (FSP) from game theory has also been adapted to train the agent against adversaries~\cite{heinrich2015fictitious}. However, neither RARL nor FSP essentially demonstrates the variations of states in an actual environment. The work closest to ours is  \cite{uesato2018rigorous} which estimates the risk using a failure search method for evaluation.

\textbf{Sampling in RL} Among methods that regulate what the RL optimizer sees, \cite{schaul2015prioritized} used the the TD error to prioritize samples in the replay buffer. Under this approach, the agent still requires to experience the scenarios, store the transition tuples, and rank them. In contrast, our approach uses predicted performance to simply reject scenarios and avoids the generation of whole trajectories. Importance sampling was used in~\cite{frank2008reinforcement} to provide an unbiased and low variance solution assuming availability of probability of rare events. Moreover, adaptive importance sampling was employed in~\cite{schmerling2016evaluating} to evaluate the probability of collision for different trajectories generated by a motion planner. Both studies assumed prior knowledge of the environment. 

\textbf{Exploration Strategy} Our method can also be viewed as an extrinsic exploration strategy. Thompson sampling \cite{kaufmann2012thompson} is commonly used in combination with RL algorithms to improve exploration. For example, \cite{haarnoja2017reinforcement} explicitly adds the policy entropy to the objective to ensure exploration. Noisy networks was used in~\cite{fortunato2017noisy} to improve exploration. Curiosity driven approaches \cite{pathak2017curiosity} are another family of exploration strategies which encourage the visitation of less visited states by giving the agent an intrinsic reward for the states with high prediction error. We believe these approaches are complimentary to our method.

\textbf{Connection with Supervised Learning} In the supervised learning framework, as training progresses it is very common to employ a mechanism to focus the training on examples that are still difficult for the model to classify \cite{galar2011review},~\cite{felzenszwalb2008discriminatively}. Such examples are either close to decision boundaries or from the less common classes. They are prioritized using their last-seen classification error. Although these ideas are simple and very effective in supervised learning, their possible usefulness in an RL context has so far been overlooked.

\section{Technical Approach}

In this section we describe the proposed method. First, we explain the architecture and training procedure for CoachNet in Section~\ref{sec:coachnet}. In Section\ref{sec:adv-sampling}, we propose a stochastic sampling approach using CoachNet's predictions to adversarially sample from the environment.

We consider the parametric Markov Decision Processes (MDP) \cite{sutton1998introduction}. An MDP is defined by a tuple $\langle S,\ A,\ R,\ T,\ \rho_0,\ \gamma \rangle$ consisting of a set of states $S$, actions $A$, a task dependent reward function $R(s,\ a)$, a transition function $T(s,\ a,\ s^\prime) = p(s^\prime|s,\ a)$, the initial state distribution $\rho_0$ and a discount factor $\gamma$. Trajectories $\tau$ are gathered according to the policy $\pi$. Given an MDP, the objective of RL is to learn a policy $\pi$ to maximize the expected sum of discounted reward $J_M(\pi) = \E_{\tau}[\sum_{t=0}^{\infty}{\gamma^t}r_t]$ where $r_t=R(s_t, a_t)$.

\subsection{CoachNet: Failure Predictor}\label{sec:coachnet}

The purpose of CoachNet is to predict the difficulty of a trajectory given the agent and its initial few interactions with the environment. In this paper, we employ failure prediction as a proxy for difficultly prediction. Failure is defined as termination (``done'') before a certain number of steps $H$ is reached.  For discussion, we call $H$ the ``prediction horizon'' corresponding to the limit that the CoachNet is able to predict given the task. 

The initial training of CoachNet follows a simple scheme. We start with the standard RL training process and let the agent learns until it reaches a certain level of performance $R_{mean}$. After that, we start collecting training data for CoachNet by storing sequences of $(s_t,\ \phi_t)$ pairs, where $s_t$ is the state and $\phi_t$ encodes information about the agent at time $t$. Each sequence of $(s_t, \phi_t)$ pairs are tagged with $c$. $c$ is 1 if the agent fails in a trajectory of length $H$ and 0 otherwise. If termination happens before $H$ steps, the rest of sequence is zero-padded. After collecting $N$ such sequences, CoachNet is trained with this data. This process is summarized in Algorithm~\ref{alg:train-1}.
   
\begin{figure}[hbt!]
	\centering
	\vspace{-10pt}
	\includegraphics[scale=0.65]{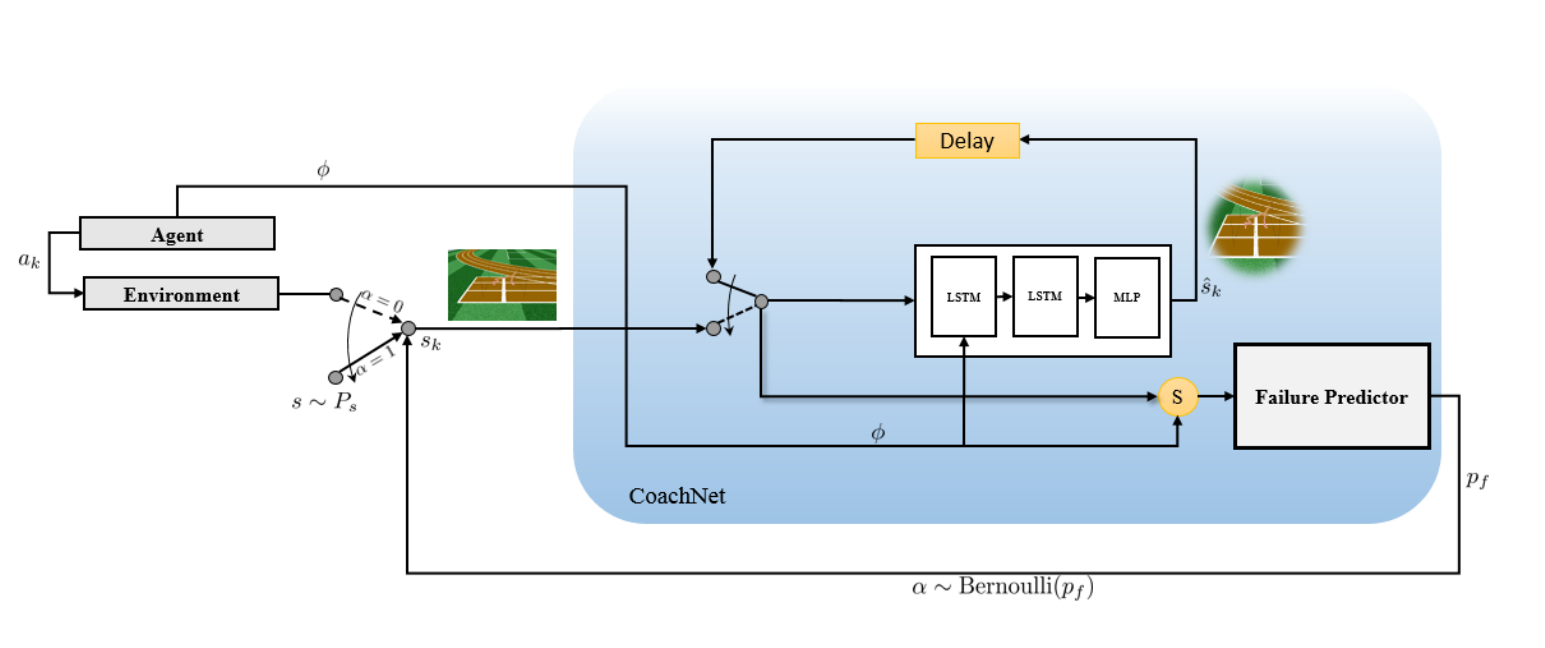}
	\caption{CoachNet Architecture. States and agent dependent parameters are fed to the state predictor. They then are stacked with predicted states and used as the inputs of feed-forward network to predict failure.}
	\vspace{-10pt}
	\label{fig:coachnet}
\end{figure}

There are many choices for the specific architecture of the CoachNet. For our main experiments, we use two cascaded components, a state predictor and a failure predictor. The state predictor is a multi-layer RNN that gets first initial $l$ state-agent pairs and predicts the states for the rest of the prediction horizon $H$. Intuitively, the state predictor encodes the state transitions given a specific agent. In experiments we show how this component improves CoachNet's ability to predict failure. After $l$ steps, the RNN uses the output of each step to predict the next state. Agent parameters, initial states, and the predicted states are then fed to a multi-layer neural network that outputs the probability of failure. The architecture of the CoachNet is illustrated in Figure\ref{fig:coachnet}.

The network is trained end-to-end by minimizing the following loss
\vspace{1pt}
\begin{align}
	L = k_1L_1(\hat{S},\ S) + k_2L_2(\hat{c},\ c)
\end{align}
\vspace{1pt}
where $L_1$ is the mean-squared error of the state prediction $\hat{S}$ and $\hat{c}$ is the predicted failure, $L_2$ is the classification loss, and $k_1$ and $k_2$ are hyperparameters.

The described training method is known as the continuation approach \cite{uesato2018rigorous}. It operates on the principle that versions of the agent from which we collect data throughout a possibly lengthy RL training process actually fail in correlated ways. The existence of such correlation makes it easier to populate failure cases in the collected data set because the agent usually fails more frequently in earlier stages of training. If failure modes of different versions of the agent were not correlate to each other, this approach could become ineffective. However, our experiments show that this is not the case in our task domains and CoachNet could decode the systematic relation between $\phi$ and $s$.

\vspace{-10pt}
\begin{algorithm}
    \caption{Stage 1 Training - Data Collection}
    \label{alg:train-1}
    \textbf{Input:} $N$ number of trajectories in first phase, $H$ prediction horizon \;
    \Repeat{$R_{mean} > threshold$}{Train the RL agent}
   	\For{n\ =\ 0\ :\ N}{
   		\For{k\ =\ 0 :\ H}{		
   			Store data $(s_t, \phi_t)$\;
   			Train the RL agent\;	
   			\If{done}{
   			Zero-pad the sequence\;
   			Tag the sequence with $c=1$ \;
   			Break\;
   			}
   		}
   		\If{not done}{
   			Tag the sequence with $c=0$ \;
   		}

    }

\end{algorithm}
\vspace{-10pt}
\subsection{Adversarial Sampling Method}\label{sec:adv-sampling}
Assuming CoachNet can be effectively trained with the setup in the previous section, we propose an adversarial sampling method that uses CoachNet's predictions to improve training over the default uniform sampling method. 

A naive approach would be letting the agent experience a scenario for the initial $l$ steps and then applying the CoachNet to make a clearcut decision about whether to continue the episode according to a certain fixed threshold or not. This naive approach has several issues. First, because the CoachNet is an estimator, there is inevitably going to be error in its predictions, which could be large especially if it has not encountered some trajectories. Second, using CoachNet in a clearcut and exclusive way could also make the agent focus on only a small subset of trajectories, making the learning process prone to over-fitting. To deal with these issues, we propose using a stochastic sampling approach. Specifically, we use the rejection sampling method \cite{bucklew2013introduction}, where we accept a trajectory with the probability of $p_f$:
\vspace{1pt}
\begin{equation}
p_f = min(f^{\alpha}(s, \phi) + \mu, 1)
\end{equation}
   
Here, $\alpha\geq0$ is a hyperparameter that is used to exponentiate the raw prediction. It can be tuned to make up for underestimation or overestimation of the failure probability prediction. $\mu$ ($0<\mu\leq1$) is another hyperparameter which ensures every trajectory has a non-zero chance of being accepted. A summary of this approach is described in Algorithm~\ref{alg:adv-sampling}.

\begin{algorithm}
    \caption{Stage 2 - Training Using Adversarial Sampling}
    \label{alg:adv-sampling}
    \textbf{Input:} $N$ number of trajectories in second phase, $T$ trajectory length, $l$ length of initial states, $M$ period for online training of CoachNet \;
   	\For{n\ =\ 0\ :\ N}{%
	   \While{not done}{
	   		\Repeat{Accepting a trajectory}{
   	   		\For{k\ =\ 0\ :\ $l$}{
   	   	 		Collect data $(s_k, \phi_k)$\;	
   	   	 	}
      		Compute $p_f$\;
      		Accept the trajectory with the probability of $p_f$\;
       	}
   	   
      \For{t\ =\ 0: T}{
      	Train the RL agent\;
      	Store data $(s_t, \phi_t)$\ for CoachNet;
      }
      \If{$(n \% M) == 0$}{
      	Fine-tune CoachNet with the new data \;
      }
  	}

     }
\end{algorithm}

\vspace{-10pt}
Once a scenario is accepted, the agent is allowed to continue the trajectory for up to an additional $T$ steps per settings of the original RL training. On the flip side, if a scenario is accurately rejected after seeing only the initial $l$ steps, it means about $T$ steps of interaction with the environment are saved. Given that $T>>l$, this could mean significantly more sample efficient training when most scenarios become simple for the agent.

Although our approach helps the agent experience difficult scenarios relatively more frequently, it also introduces a bias to the sampling in contrast to vanilla Monte Carlo sampling. We use a simple but practical approach to handle such a bias. Typical RL process is significantly non-stationary because of the changing policy and state distributions and thus unbiased updates is comparably much more important closer to convergence or to the end of training. Hence we use a schedule on $\mu$ to modulate the bias by gradually increase $\mu$ from an initial value $\mu_0$ to 1 to allow a smooth transition to vanilla Monte Carlo sampling. Note that this schedule is well aligned with the proposed system. As we train the agent with more adversarial samples and train CoachNet in an online fashion, the agent gets more robust and the probability of failure gets closer to 0 for most of the scenarios.
\footnote{An alternative way of dealing with this bias is to use importance sampling \cite{bucklew2013introduction} weights in the RL optimization step. This approach requires computing failure probability for all sampled transitions, which could be very difficult in continuous cases and in general computationally expensive. A potentially better approach is to estimate failure probability density ratios of whole trajectories solely based on the initial $l$ states. This is a potential topic for further research.}

Yet another important aspect of our design is the continued fine-tuning of CoachNet with new data every M iterations. In this step we use all collected data in both stages. However, similar to the initial training of CoachNet the recent data is sampled with higher probability. As noted above, this helps keeping the CoachNet up to date as the agent's state visitation distribution changes along with the agent's policy.

\section{Experimental Results}
\begin{figure}[hbt!]
	\centering
	\includegraphics[scale=0.5]{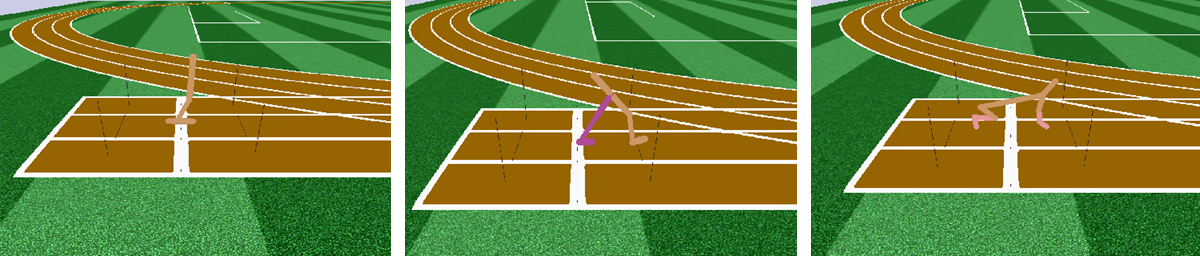}
	\caption{\textbf{Continuous Control Environments:} left-to-right: the RoboschoolHopper, RoboschoolWalker2D, RobochoolHalfCheetah}
	\label{fig:envs}
\end{figure}

We evaluate the performance of the proposed adversarial sampling method on three continuous control environments in Roboschool \cite{schulman2017proximal} including RoboschoolHopper-v1, RoboschoolHalfCheetah-v1, and RoboschoolWalker2D-v1, as shown in Figure~\ref{fig:envs}. We used the forward walker reward to keep things simple. 
While the proposed method does not make any assumptions on the RL algorithm used for training the agent and can be applied to any RL pipeline, we build our algorithm on Proximal Policy Optimization (PPO) \cite{schulman2017proximal} for the sake of evaluation in this section. PPO was selected for RL training as it has demonstrated reasonable performance across environments used in this paper. The objective is to investigate the effect of the CoachNet guided sampling method on the performance of the chosen RL algorithm. 

\vspace{-10pt}
\subsection{CoachNet}
In this section, we asses the performance of CoachNet in predicting agent's failure in different environments. We also compare the proposed architecture to a feed-forward network which does not have the state-predictor component. Failures are defined by the standard termination criteria of each environment so as to make the experiments reproducible. 

For each environment, the agent was trained for 200k timesteps. A publicly available implementation of PPO \cite{stable-baselines} was employed, and the default parameters provided in \cite{schulman2017proximal} was used for agent training. To train CoachNet, we used the same data used for training of the RL agent. We trained the agent until it reached to a certain performance in terms of the average episode reward, after which we started collecting data for training CoachNet. As mentioned in Section 3.1, in order to get more samples from the failure cases, we used the continuation approach to train CoachNet. Also, since the number of failed episodes is less than the number of successful episodes, we used a subset of data to avoid the imbalanced training data problem. To this end, the collected data was sampled using two criteria, giving more priority to more recent data and keeping the ratio of the number of failure cases to the total number of scenarios between 0.3 and 0.7.

Five initial states were fed as inputs to CoachNet, and the agent's age was used as the agent-dependent parameter. 

We investigated two sets of architectures for CoachNet; one with state-predictor and one without state-predictor, hereafter referred to as CoachNet-WSP and CoachNet-MLP, respectively. For state prediction in CoachNet-WSP a two-layer LSTM \cite{gers1999learning} with [1024, 512] units and a one layer neural network with the same number of units as the environment's states was used in an RNN cell. A multi-layer feed-forward network was also adopted for failure predictor component in both CoachNet-WSP and CoachNet-MLP architectures. For CoachNet-WSP we used layers with [1024, 512] units and for CoachNet-MLP the network has [1024, 512, 256, 128] units. We tried different settings and chose the one with better performance. For example, increasing the capacity of CoachNet-MLP resulted in over-fitting so we avoided bigger networks.
 
Table~\ref{table:fspresults} summarizes the evaluation result of CoachNet on 100 seeds. Instead of precision and recall, which are commonly used in supervised learning, we report the average probability of failure and success for successful and failed trajectories on all of the environments as these numbers directly affect the sampling performance. As mentioned in Section\ref{sec:adv-sampling} the probability of accepting a trajectory is a function of the probability of failure predicted by CoachNet.
The result shows that CoachNet-WSP outperforms the CoachNet-MLP architecture in all of the locomotion tasks. We believe that the role of the state predictor becomes more crucial when dealing with more dynamic environments as it can better model the agent's dynamics.

\begin{table*}
	\footnotesize
	\begin{center}
		\begin{tabular}{cc||cccc||cccc||cccc}
			&& \multicolumn{4}{c||}{RoboschoolHopper-v1} & \multicolumn{4}{c||}{ RoboschoolHalfCheetah-v1}& \multicolumn{4}{c}{RoboschoolWalker2D-v1}\\
			\cline{3-14}			
			\multicolumn{2}{c||}{Algorithms} & \multicolumn{2}{c}{Failure} & \multicolumn{2}{c||}{Success} & \multicolumn{2}{c}{Failure} & \multicolumn{2}{c||}{Success} & \multicolumn{2}{c}{Failure} & \multicolumn{2}{c}{Success} \\
			\hline
			\multicolumn{2}{c||}{CoachNet-WSP} & \multicolumn{2}{c}{0.88} & \multicolumn{2}{c||}{0.19} & \multicolumn{2}{c}{0.97} & \multicolumn{2}{c||}{0.14} & \multicolumn{2}{c}{0.84} & \multicolumn{2}{c}{0.25} \\
			\multicolumn{2}{c||}{CoachNet-MLP} & \multicolumn{2}{c}{0.65} & \multicolumn{2}{c||}{0.41} & \multicolumn{2}{c}{0.73} & \multicolumn{2}{c||}{0.32} & \multicolumn{2}{c}{0.57} & \multicolumn{2}{c}{0.49} \\
			
		\end{tabular}
	\end{center}
	\vspace{-5pt}
\caption{Results of failure prediction by CoachNet-WSP and CoachNet-MLP tested on 100 seeds. The average probability of failure is reported on both failed and successful scenarios. In both of these scenarios CoachNet-WSP has a better performance in all environments.}
\label{table:fspresults}
	\vspace{-10pt}
\end{table*}

\subsection{Adversarial Sampling Method}
In this section, we evaluate the impact of the proposed adversarial sampling method on the performance of the RL algorithm. We trained two sets of agents; one uses the conventional vanilla Monte Carlo sampling method and another one uses the proposed adversarial approach. For this experiment, we restored the weights of the trained agents in the previous section as initialization to resume the training. We also fine tuned CoachNet after every 400k timesteps. We found this step to have a minor effect on the agent's performance. However, we believe that the effect will be more tangible if the task is more difficult and if the behavior of the agent changes more significantly during the second phase of the optimization.

\begin{figure}[b!]
	\centering
	\vspace{-20pt}
	\includegraphics[scale=0.3]{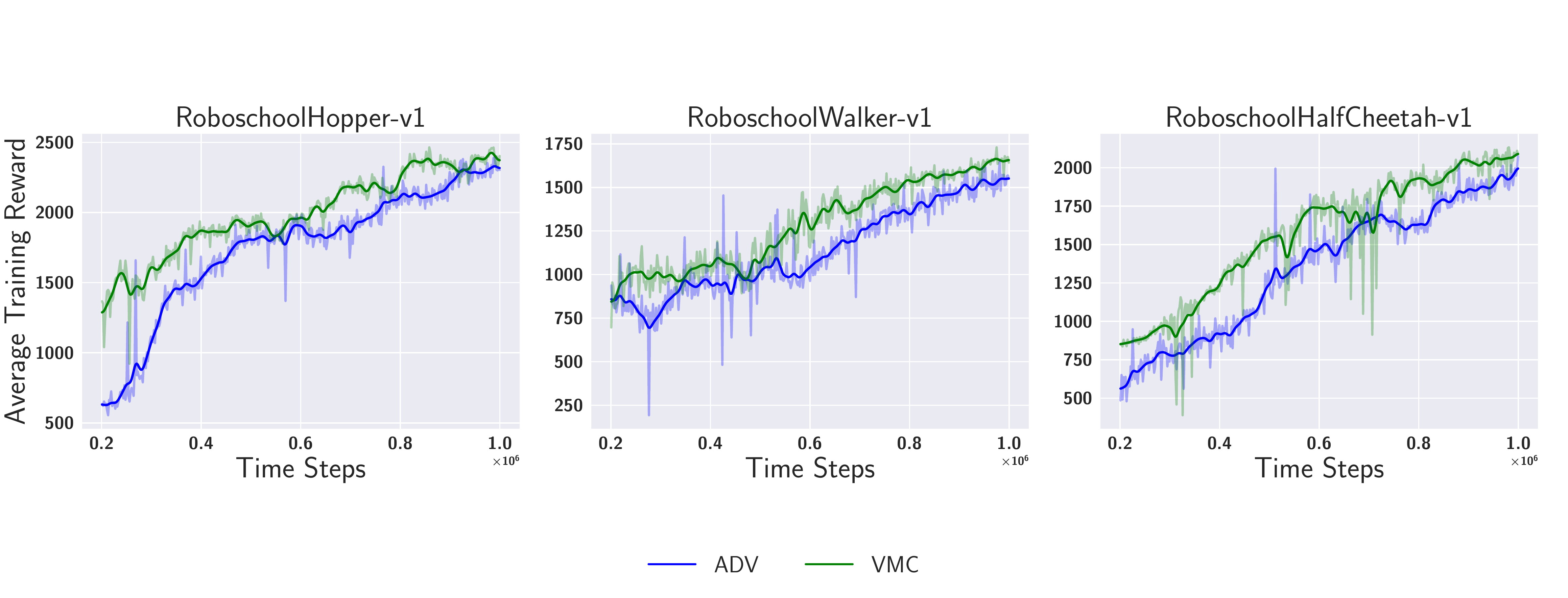}
	\vspace{-20pt}
	\caption{Comparison of the training time average reward between vanilla Monte Carlo (VMC) vs. Adversarial Sampling (ADV) .}
	\label{fig:training-rew}
	\vspace{-15pt}
\end{figure}

We trained both agents on one million time steps. The average episode reward is shown in Figure~\ref{fig:training-rew}. In all of the locomotion tasks, training with adversarial sampling approach has resulted in lower average reward during the training. This behavior is expected as CoachNet intentionally guides the agent to harder scenarios. The difference become smaller as the agent gets trained on the adversarial trajectories. 

\begin{figure}[hbt!]
	\centering
	\vspace{-20pt}
	\includegraphics[scale=0.3]{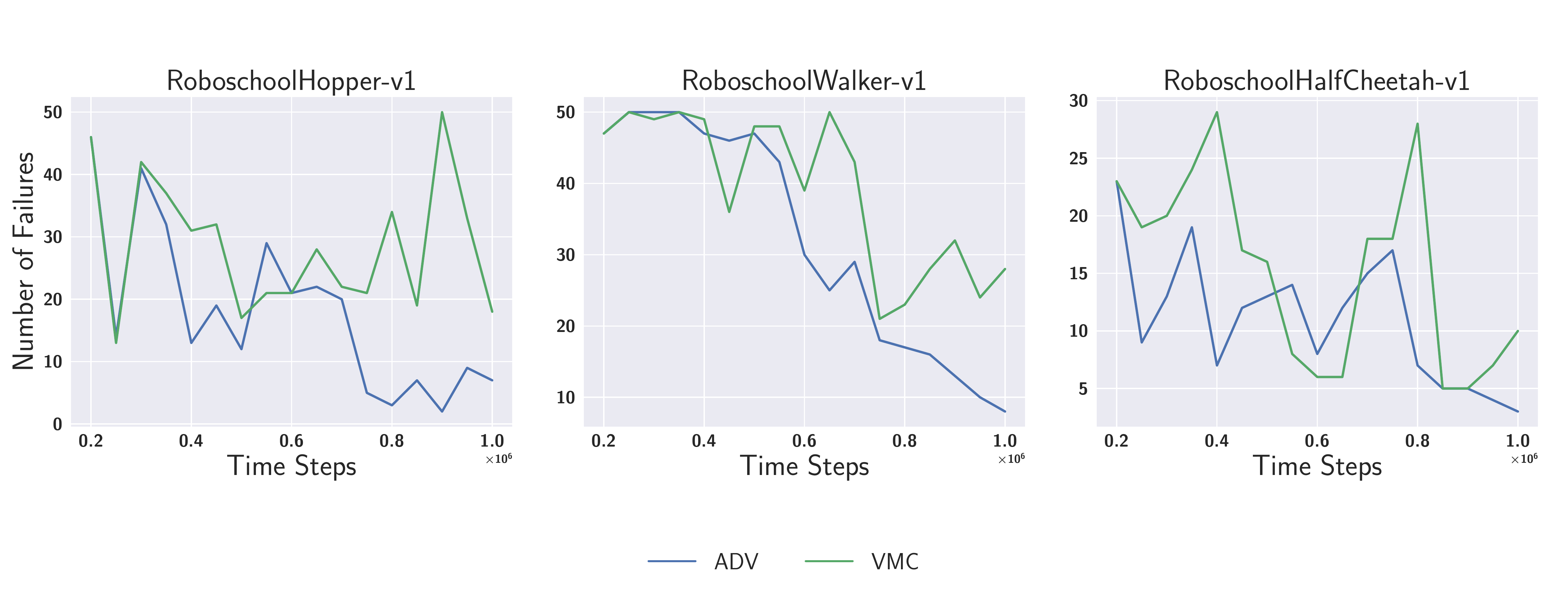}
	\vspace{-20pt}
	\caption{Comparison of the test-time average reward between VMC vs.ADV. Tested with 50 random seeds.}
	\label{fig:eval-fail}
	\vspace{-15pt}
\end{figure}

\begin{figure}[hbt!]
	\centering
	\vspace{-10pt}
	\includegraphics[scale=0.3]{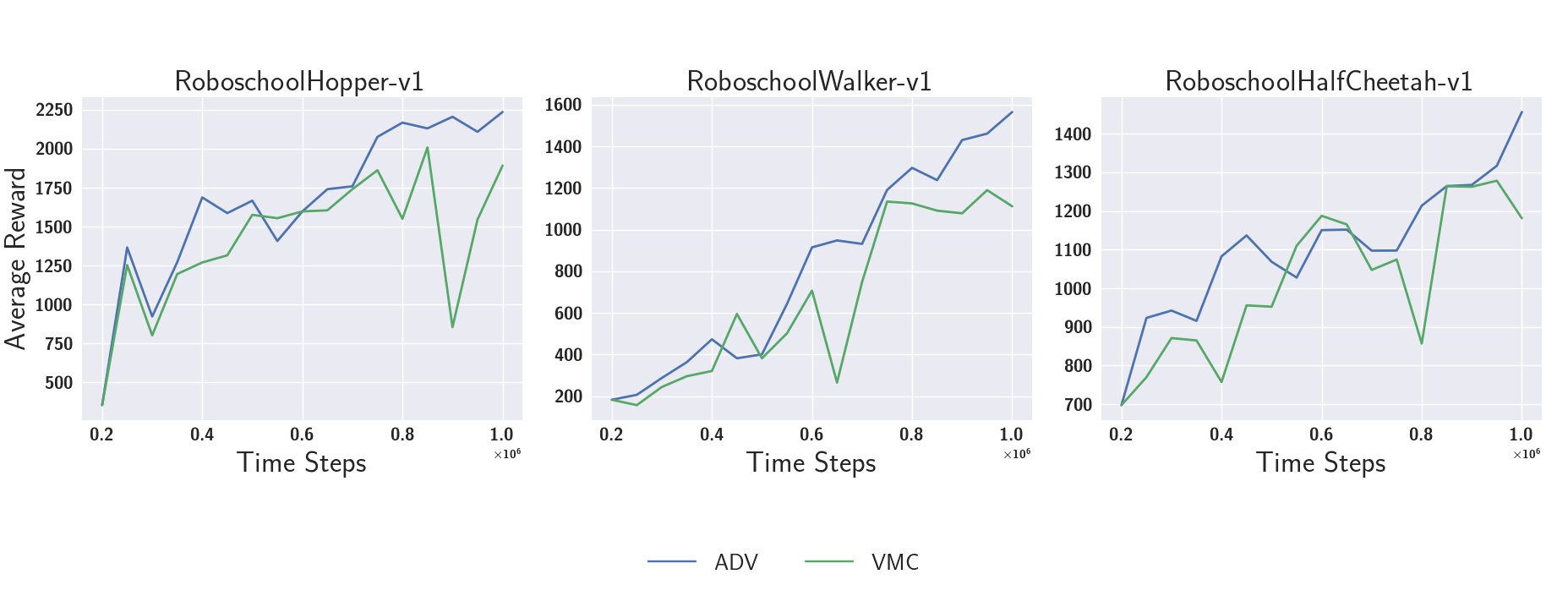}
	\vspace{-20pt}
	\caption{Comparison of the test-time average episode reward between VMC vs.ADV. Tested with 50 random seeds.}
	\label{fig:eval-rew}
	\vspace{-15pt}
\end{figure}

In order to evaluate the performance of the agents trained with these two approaches, we stored the agent model in every 50k timesteps. We then evaluate each agent on 50 seeds. Figures~\ref{fig:eval-fail}~and~\ref{fig:training-rew} show the average number of failures and episode reward for each agent, respectively. As shown in Figure~\ref{fig:eval-fail}, the number of failures are almost halved for agents trained using the adversarial approach compared to those of agents trained using VMC sampling method. Figure~\ref{fig:eval-rew} also shows the similar result in all of three environments. As can be seen from the figure, the average reward of the evaluation is higher after 500k timesteps for the agents trained using the proposed method.

\section{Conclusion}

In this paper we introduced CoachNet, where a predictor is first trained to predict the difficulty of scenarios for the RL agent and such predictions are in turn used to guide sampling for RL training to focus on the more challenging trajectories. We showed that agents thus trained for an equal number of time steps achieve higher average reward and lower number of failures at test time in comparison to agents trained with the vanilla Monte Carlo sampling method.

While we studied the effect of CoachNet on the performance of PPO, it should be obvious that the CoachNet method can be combined with other RL algorithms as well. For example, a transition-level predictor (as versus trajectory-level predictor) can be similarly constructed to prioritize samples in a replay buffer used by many RL algorithms. In addition, while we used CoachNet in an adversarial sampling setting, it can also be used for online monitoring and risk estimation.

There are also various ways of extending the CoachNet architecture. We are currently working on using CoachNet to generate adversarial scenarios for potentially even more efficient training of RL agents with robust performance. Moreover, failures come in different kinds, an autonomous car or a service robot may indeed have to differentiate crashes that destroys itself and crashes that also kill humans. While we treated CoachNet as predictor of undifferentiated failure, we can extend it to admit multiple classes of failures so as to enable intelligent trade-offs among different types of failures as well as with reward accumulation. 

The specific training setup for this first version of CoachNet exploits the strong correlation between initial states and final outcome where the episodes are short. However, this basic idea of CoachNet can be generalized to long episodes and continuous cases so long as a relevant temporal horizon can be established. In autonomous driving, for example, even though going from A to B may take hours, the temporal horizon for safety-critical decision typically does not exceed 10 seconds. Robotic manipulation and robot locomotion have similar temporal regularity that can be exploited. More generally, where highly variable temporal horizons are relevant for the same task, we could employ multiple concurrent predictors trained with RL-style learning algorithm~\cite{sutton2011horde} to make relevant predictions based on a short sequence of steps happening at any time, rather than just at the beginning of short episodes.

\bibliographystyle{IEEEtran}
\bibliography{ref}  

\end{document}